\documentclass[iicol]{sn-jnl}
\usepackage{graphicx} 
\usepackage{tikz} 
\usepackage{listings}
\usepackage{enumitem}
\usepackage{graphicx}
\usepackage{multicol, latexsym}
\usepackage{blindtext}
\usepackage{caption}
\usepackage{longtable}
\usepackage{csquotes}
\usepackage{amsfonts}
\usepackage{amsmath}
\usepackage{amsthm}
\usepackage{amssymb}
\usepackage{algorithm}
\usepackage{tabularx}
\usepackage{capt-of,lipsum} 
\usepackage{balance}
\usepackage{subfigure} 
\usepackage{listings}
\usepackage{comment}
\usepackage{diagbox}
\usepackage{pdflscape}
\usepackage{booktabs}
\usepackage{makecell}
\usepackage{lipsum}
\usepackage{adjustbox}
\usepackage{float}
\usepackage{makecell}
\usepackage{graphicx}%
\usepackage{multirow}%
\usepackage{amsmath,amssymb,amsfonts}%
\usepackage{amsthm}%
\usepackage{mathrsfs}%
\usepackage[title]{appendix}%
\usepackage{xcolor}%
\usepackage{textcomp}%
\usepackage{manyfoot}%
\usepackage{booktabs}%
\usepackage{algorithm}%
\usepackage{algorithmicx}%
\usepackage{algpseudocode}%
\usepackage{listings}%
\usepackage{svg}

\newcommand\2{\textit{(ii)}}

\newcommand{\blue}[1]{\textcolor{black}{#1}}

\begin{document}
\title[An unsupervised decision-support framework for multivariate biomarker analysis in athlete monitoring
]{An unsupervised decision-support framework for multivariate biomarker analysis in athlete monitoring
}
\author*[1]{\fnm{Fernando} \sur{Barcelos Rosito}}\email{fernando.rosito@ufcspa.edu.br}

\author[2]{\fnm{Sebastião} \sur{De Jesus Menezes}}\email{sebastiaojpp@gmail.com}

\author[2]{\fnm{Simone} \sur{Ferreira Sturza}}\email{simonesturza@hotmail.com}

\author[1]{\fnm{Adriana} \sur{Seixas}}\email{adrianaseixas@ufcspa.edu.br}

\author*[1]{\fnm{Muriel} \sur{Figueredo Franco}}\email{muriel.franco@ufcspa.edu.br}

\affil*[1]{\orgname{Federal University of Health Sciences of Porto Alegre (UFCSPA),}
\orgaddress{\state{RS}, \country{Brazil}}}

\affil[2]{\orgname{Levino Inova}, \orgaddress{\state{AC}, \country{Brazil}}}


\abstract{\textbf{Purpose.} Athlete monitoring in practice is constrained by small cohorts, heterogeneous biomarker scales, limited feasibility of repeated biological sampling, and the lack of reliable injury ground truth. These limitations reduce the interpretability and utility of traditional univariate and binary risk models. This study aims to address these challenges by proposing an unsupervised multivariate framework to identify latent physiological states in athletes using real data.
\textbf{Methods}. We propose a modular computational framework that operates directly in the joint biomarker space and integrates data preprocessing with clinical safety screening, unsupervised clustering, and centroid-based physiological interpretation. Physiological profiles are learned exclusively from data collected from amateur soccer players during a competitive microcycle. Synthetic data augmentation is applied to evaluate robustness and scalability. Ward hierarchical clustering is used for monitoring and etiological differentiation, while Gaussian Mixture Model (GMM) augmentation supports structural stability analysis in high-dimensional settings.
\textbf{Results.} The framework identifies physiologically coherent profiles that distinguish mechanical damage from metabolic stress while preserving the dominance of homeostatic states. Results from synthetic data augmentation demonstrate the framework's feasibility and ability to detect latent, silent risk phenotypes that are typically missed by conventional univariate monitoring. Structural stability analyses indicate that the framework remains robust under data augmentation and higher-dimensional settings.
\textbf{Conclusion}. The proposed framework enables interpretable identification of latent physiological states from multivariate biomarker data without reliance on injury labels. By distinguishing underlying physiological mechanisms and revealing silent risk patterns not captured by conventional monitoring, it provides actionable insights to support clinicians, physiologists, and sports health professionals in individualized athlete monitoring and informed decision making.}

\maketitle

\section{Introduction}
In elite sports, maximizing performance ultimately comes down to managing internal load. This means, for example, understanding how the body responds to external stressors of training and the costs it incurs for specific performance indicators \cite{Contreras2024}. This internal response represents a dynamic physiological adjustment as the body strives to maintain homeostasis under physical demand \cite{Haller2023}. When the magnitude of training stress exceeds the athlete’s capacity to recover, it can trigger maladaptive states (\textit{e.g.}, non-functional overreaching and chronic fatigue-like syndromes) that compromise both health and long-term athletic development \cite{LopezCuervo2025}.

Historically, reliance has been placed on isolated snapshots of biomarkers (\textit{e.g.}, Creatine Kinase (CK), C-reactive protein (CRP), and Cortisol) to track the cost of stress of training and external factors. However, because human physiology operates as a complex, non-linear network, univariate analysis is often insufficient. Current approaches often fail to capture the athlete's global state, and critical interactions between biological systems are overlooked \cite{saidi2021hematology}. Although CK, for example, is frequently used as an indicator of exercise-induced muscle damage, isolated elevations provide limited information about the athlete’s overall physiological state. As discussed in the literature, the absence of concurrent hormonal or inflammatory data makes it difficult to distinguish between normal training adaptation and maladaptive fatigue processes \cite{Armstrong2022}.

New laboratory technologies have inadvertently introduced a new challenge, the curse of dimensionality \cite{berisha2021digital}. Although access to dozens of biomarkers is now available, the volume of data far exceeds human processing capacity. This paradoxically leads back to analytical reductionism. Consequently, recent literature has turned toward Machine Learning (ML) to handle this complexity \cite{VanEetvelde2021}. There are works, for instance, that establish a strong benchmark by showing that integrating hematological profiles with Global Positioning System (GPS) data significantly improves injury prediction \cite{Rossi2023}. 

However, current approaches have a limitation: they rely mostly on binary clustering, where risks are defined only as low or high. While useful, this offers low prescriptive value \cite{Losciale2024,VanEetvelde2021} by simply flagging an athlete as \textit{at risk}. This does not elucidate whether the root cause is, for example, a mechanical or metabolic issue, such as tissue damage or neuroendocrine suppression \cite{Armstrong2022,Leckey2024}. Different conditions like those require adequate statistical and data analysis for data-informed decisions \cite{Martin2022}. This can result in tailored recovery strategies and a better understanding of the biological underpinnings of athletes' global states. For this, it is essential to have novel, automated tools that simplify the processing, analysis, and visualization of biomarker correlations \cite{VanEetvelde2021,Leckey2024}.

A further limitation in this field concerns data scarcity, which remains a structural constraint in elite sports research. Due to the logistical, financial, and ethical challenges associated with repeated biological sampling, many benchmark studies rely on very small cohorts. For instance, widely cited works on injury risk and workload monitoring have used sample sizes on the order of one professional squad, often involving fewer than 20 athletes \cite{Rossi2023,Foucaud2025}. While these studies provide high-quality and well-controlled datasets, the limited number of observations constrains statistical generalization and hampers the assessment of model scalability, particularly for multivariate and unsupervised learning approaches.


In this work, we bridge these gaps by proposing a computational framework based on unsupervised multivariate analysis of biomarker data. Rather than framing athlete monitoring as a binary risk classification task, the pipeline applies clustering directly to the joint biomarker space, allowing multiple physiological states to emerge from the data. By increasing the clustering resolution to $k = 5$, the analysis enables a finer characterization of biological profiles associated with distinct underlying stress mechanisms. To evaluate the robustness of this clustering strategy under data regimes exceeding the size of typical elite sports cohorts, Gaussian Mixture Models (GMMs) are employed to model the joint distribution of observed biomarkers and generate synthetic data. This framework allows for scalability analysis and the assessment of cluster stability in simulated large-scale scenarios derived from real-world measurements. Beyond exploratory analysis, the proposed framework is explicitly designed to serve as a decision-support component within physiological monitoring systems.

The key contributions of this article can be summarized as follows:
\begin{itemize}
\item A multi-level computational pipeline for athlete monitoring that enables both macro-level assessment through $k=3$ and finer etiological characterization via $k=5$;
\item The application of Gaussian Mixture Models (GMM) for synthetic data generation and scalability validation in small-sample sports biomarker datasets;
\item The analysis and differentiation between mechanical damage and metabolic stress through unsupervised learning, providing higher prescriptive value for recovery interventions; and
\item The validation of algorithmic sensitivity through the isolation of synthetically injected silent risk patterns, demonstrating the potential to identify multivariate anomalies overlooked by traditional analysis.
\end{itemize}

The rest of this paper is organized as follows. Section 2 discusses related work regarding biomarker monitoring and the application of machine learning in elite sports. Section 3 details the proposed methods, encompassing the research design, data architecture, and the generative simulation workflow. Section 4 presents the experimental results, including the algorithmic validation on real-world data and the physiological characterization of the identified clusters. Section 5 discusses overcoming the traditional risk dichotomy and the validity of using GMM in small-data scenarios in sports science. Finally, Section 6 concludes the article and provides details on future work.
\section{Related Work}
Blood-based biomarkers have become a cornerstone for monitoring physiological stress, fatigue, and recovery in athletes \cite{pedlar2019blood}. Several systematic reviews have established that markers such as CK, cortisol, and CRP provide valuable insights into muscle damage, inflammation, and endocrine responses to training load \cite{saidi2021hematology, Haller2023, Contreras2024}. These biomarkers reflect the body's adaptive response to external stressors, which emerges from complex interactions across multiple physiological systems \cite{Armstrong2022, LopezCuervo2025}. However, individual biomarkers often exhibit high interindividual variability and limited specificity, making it difficult to distinguish between adaptive physiological responses and maladaptive stress when analyzed in isolation \cite{Souglis2024}.

The multidimensional nature of physiological adaptation has motivated the application of multivariate analysis and unsupervised learning methods to identify latent physiological profiles \cite{gonzalez2025unsupervised}. Clustering techniques, including hierarchical clustering and K-Means, have been successfully applied to characterize athlete physiological states and identify patterns associated with injury risk and recovery \cite{Martin2022, Popczyk2025}. For example, Foucaud et al. demonstrated that unsupervised learning can reveal distinct recovery patterns across an Olympic training cycle, highlighting the ability of clustering approaches to capture meaningful physiological dynamics without relying on predefined labels \cite{Foucaud2025}. These approaches provide a more holistic representation of physiological state compared to reductionist univariate monitoring.

Beyond exploratory profiling, machine learning methods have also been applied for injury prediction and risk assessment. Systematic reviews have shown that supervised models, including Random Forest and gradient boosting, can achieve promising performance in predicting injury risk from physiological and workload data \cite{VanEetvelde2021, Leckey2024}. In this direction, Rossi et al. demonstrated that incorporating biomarker-derived physiological profiles into predictive models significantly improved injury forecasting accuracy in elite soccer players \cite{Rossi2023}. However, these approaches depend heavily on labeled injury data, which are often scarce, noisy, or unavailable, limiting their applicability in real-world monitoring scenarios.

\begin{table*}[ht]
\centering
\caption{Comparison of existing approaches for athlete biomarker analysis and physiological modeling.}
\label{tab:comparison}
\begin{tabular}{ccccccc}
\toprule
\textbf{Study} &
\textbf{Year} &
\textbf{Multivariate} &
\textbf{Unsupervised} &
\textbf{\makecell{Latent \\ States}} &
\textbf{Interpretable} &
\textbf{\makecell{Small Data \\ Robust}}
\\
\midrule
\cite{pedlar2019blood} & 2019 
& Partial & No & No & Yes & Yes \\
 \cite{Martin2022} & 2022
& Yes & Yes & Partial & Partial & Partial \\
 \cite{Rossi2023} & 2023
& Yes & Partial & Partial & Partial & No \\
\cite{Leckey2024} & 2024
& Yes & No & No & No & No \\
\cite{gonzalez2025unsupervised} & 2025
& Yes & Yes & Yes & Partial & No \\

\cite{Foucaud2025} & 2025
& Yes & Yes & Partial & Partial & Partial \\
\makecell{This Work} & 2026
& Yes
& Yes
& Yes
& Yes
& Yes

\\
\bottomrule
\end{tabular}
\end{table*}

Another key limitation concerns the widespread use of binary risk classification frameworks that categorize athletes as high- or low-risk based on predefined thresholds. Clinical and methodological analyses have shown that such dichotomous classification oversimplifies the underlying physiological complexity and may obscure important etiological distinctions between different stress mechanisms \cite{Losciale2024}. Moreover, the increasing availability of high-dimensional physiological data introduces additional analytical challenges, including the curse of dimensionality, which complicates statistical modeling and interpretation \cite{berisha2021digital, Hair2019}.

Recent research has also explored the use of generative modeling techniques to address limitations associated with small sample sizes and data scarcity in biomedical and sports contexts \cite{van2024synthetic}. For example, synthetic data generation using probabilistic models has emerged as a promising approach to improve model robustness and enable scalable analysis while preserving the underlying statistical structure \cite{waseem2025review}.

Despite these advances, important gaps remain. Most existing studies either rely on supervised prediction models that require injury labels or use unsupervised clustering primarily as a preprocessing step for predictive tasks. Few works focus on developing interpretable, unsupervised frameworks specifically designed to identify latent physiological states directly from multivariate biomarker structure, while addressing the constraints of small sample sizes and high dimensionality. A comparison of existing solutions in the literature is provided in Table \ref{tab:comparison}, highlighting the opportunities and contributions this work offers to the field of biomarker correlation.

To address these limitations, this study proposes a fully unsupervised computational framework that integrates multivariate clustering, physiological interpretation, and generative validation. By operating directly in the joint biomarker space without reliance on injury labels, the proposed approach enables robust identification of interpretable physiological states and provides a scalable decision-support framework for athlete monitoring.

It is important to mention that the key distinction of the proposed framework lies not merely in applying clustering to biomarker data, but in integrating unsupervised physiological state discovery, interpretability via centroid analysis, and structural validation under small-sample constraints into a unified decision-support architecture designed for real-world health monitoring systems. Unlike prior studies that use clustering primarily as a preprocessing step or exploratory tool, the proposed framework operationalizes clustering as the core decision-support mechanism.
\section{Approach}

We define an unsupervised and multi-stage computational framework for physiological state identification in athletes based on multivariate analysis of blood biomarkers. The proposed framework was explicitly designed to operate under real-world sports monitoring constraints, including limited cohort sizes, heterogeneous biomarker scales, and the absence of reliable injury ground truth.

Rather than framing athlete monitoring as a supervised injury-prediction task, the framework focuses on discovering latent physiological profiles directly from the joint structure of biomarker data. In this framework, physiological profiles are learned exclusively from real-world data, while synthetic data are employed solely for robustness, scalability, and structural validation. 

\begin{figure*}[ht]
    \centering
    \includegraphics[width=\textwidth]{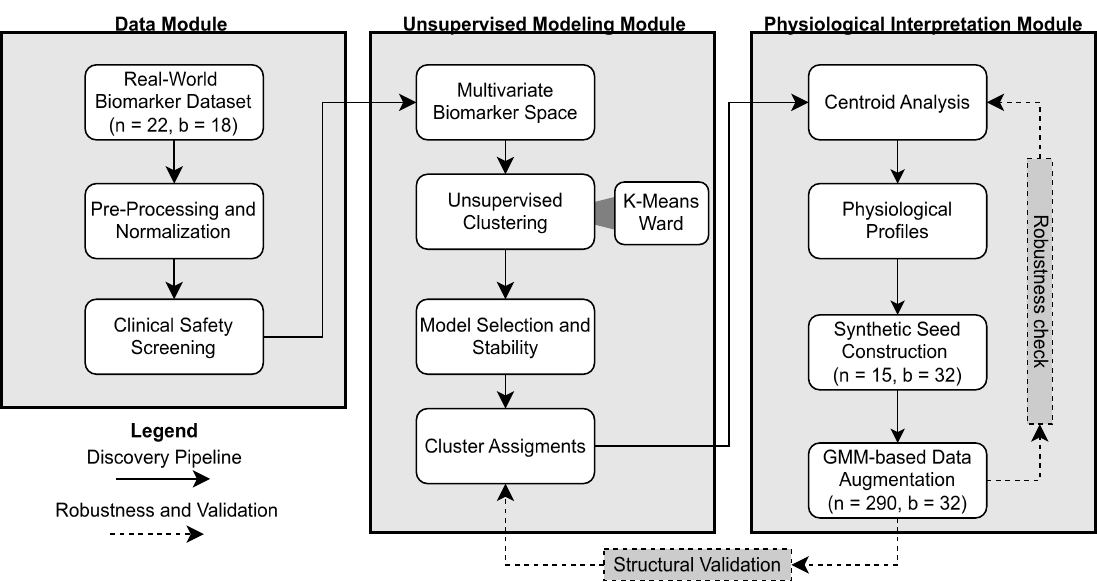}
    \caption{Overview of the Proposed Framework, where $n$ denotes the number of athletes and $b$ the number of available biomarkers}
    \label{fig:approach}
\end{figure*}

As summarized in Figure~\ref{fig:approach}, the framework is organized into (\textit{i}) a \textit{Data Module} for acquisition and preprocessing, (\textit{ii}) an \textit{Unsupervised Modeling Module} for clustering and model selection, and (\textit{iii}) a \textit{Physiological Interpretation Module} for centroid-based profiling, with an auxiliary validation path for robustness and scalability.

Each module is responsible for specific tasks that comprise the framework proposed. We also defined two different pipelines: one for knowledge discovery (solid lines) and one for structural validation and robustness check (dashed lines). As shown, the robustness check occurs only in the Physiological Interpretation Module, whereas Structural Validation involves interactions in the Unsupervised Modeling Module. However, both rely on the GMM-based data augmentation approach applied. The framework and the tasks that compose each module are detailed in the rest of this section.

\subsection{Real-World Dataset and Physiological Space Definition}

The baseline dataset used in this study was derived from an anonymous, real-world dataset comprising 22 male amateur soccer players in the western region of Northern Brazil (mean age: $24.5 \pm 3.2$ years; training frequency: 2--3 sessions per week). The dataset was initially collected by a startup specializing in biomarker correlation. Data acquisition followed a standard competitive microcycle monitoring protocol, with blood sampling conducted at three time windows: Pre-Match (baseline), immediately Post-Match (0h), and Recovery (24h).  \blue{The dataset used in this study contains anonymized physiological measurements and does not include any personally identifiable information. All methods were carried out in accordance with relevant guidelines and regulations, and informed consent was obtained from all participants whenever required. The study protocol was approved by the Institutional Research Ethics Committee of União Educacional do Norte (UNINORTE) under the number 5.159.860}.

Eight biomarkers spanning enzymatic, hormonal, and inflammatory domains were analyzed across different time windows, yielding 18 combinations, as detailed in Table~\ref{table:biomarkers}. This real-world dataset ($n = 22$) defines the physiological space from which latent profiles are discovered and serves as the exclusive source for learning cluster structure.

\begin{table*}[ht]
\centering
\caption{Biomarkers used for multivariate physiological profiling and their acquisition windows}
\label{table:biomarkers}
\begin{tabular}{c c c}
\toprule
\textbf{Biomarker} & \textbf{Description} & \textbf{\makecell{Time \\ Windows}} \\
\midrule
Creatine Kinase (CK) &
\makecell[l]{Enzymatic marker of skeletal muscle\\damage and mechanical stress} &
\makecell{Pre\\Post\\24h} \\

Lactate Dehydrogenase (LDH) &
\makecell[l]{Enzyme associated with tissue breakdown\\and metabolic stress} &
\makecell{Pre\\Post\\24h} \\

C-Reactive Protein (CRP) &
\makecell[l]{Acute-phase inflammatory marker\\reflecting systemic inflammation} &
\makecell{Pre\\Post\\24h} \\

Cortisol &
\makecell[l]{Hormonal marker of hypothalamic--\\pituitary--adrenal axis activation\\and physiological stress} &
\makecell{Pre\\Post\\24h} \\

Total Testosterone &
\makecell[l]{Anabolic hormone associated with\\recovery status and training adaptation} &
\makecell{Pre\\Post\\24h} \\

Peripheral Oxygen Saturation (SpO\textsubscript{2}) &
\makecell[l]{Indicator of peripheral oxygenation\\and cardiorespiratory status} &
\makecell{Pre} \\

Resting Heart Rate &
\makecell[l]{Cardiovascular marker reflecting\\autonomic balance and fatigue} &
\makecell{Pre} \\

Arterial Blood Pressure &
\makecell[l]{Hemodynamic indicator of cardiovascular\\load and systemic response} &
\makecell{Pre} \\
\bottomrule
\end{tabular}
\end{table*}

\subsection{Pre-processing and Clinical Safety Screening}

Given the scalar heterogeneity across biomarkers (\textit{e.g.}, creatine kinase on the order of $10^3$ U/L versus C-reactive protein on the order of $10^0$ mg/L), Z-score normalization was applied to equalize variable weighting in multivariate distance calculations. This is shown in Equation \ref{eq1}, where $x_{ij}$ represents the value of biomarker $j$ for athlete $i$, and $\mu_j$ and $\sigma_j$ denote the mean and standard deviation of biomarker $j$, respectively.

\begin{equation} 
z_{ij} = \frac{x_{ij} - \mu_j}{\sigma_j}
\label{eq1}
\end{equation}

To ensure clinical safety and data integrity, multivariate outlier detection was performed using Euclidean distance (Equation \ref{eq2}) in the normalized biomarker space. Subjects exhibiting distances greater than 25 units from the global centroid were flagged as clinical outliers, indicating potential measurement errors or acute pathological conditions, and were excluded from the clustering process.

\begin{equation} 
d(x, y) = \sqrt{\sum_{j=1}^{p} (x_j - y_j)^2}
\label{eq2}
\end{equation}

\subsection{Unsupervised Clustering and Algorithm Selection}
Latent physiological profiles were identified using unsupervised clustering applied exclusively to the real-world dataset. Two clustering techniques were evaluated: K-Means and Agglomerative Hierarchical Clustering with Ward linkage. K-Means was employed as a baseline comparator due to its widespread use and computational simplicity. However, given its sensitivity to initialization and reduced stability in small-sample settings, it was not selected as the primary clustering method.

Ward’s hierarchical clustering was adopted as the main algorithm due to its robustness in heterogeneous and small datasets, as supported by prior literature \cite{Popczyk2025}. This method minimizes within-cluster variance while preserving hierarchical relationships between subjects.

The determination of the optimal number of clusters ($k$) was not arbitrary. Silhouette scores were computed to assess cluster cohesion and separation, while dendrograms were inspected to evaluate hierarchical structure. Stability was rigorously tested by executing each configuration ten times with distinct random seeds. Two resolutions were retained: $k=3$ for macro-level physiological monitoring and $k=5$ for fine-grained etiological differentiation.

To facilitate physiological interpretation, heatmaps were generated using normalized centroid values (Z-scores), enabling direct comparison of biomarker magnitudes across heterogeneous units on a standardized scale.

Following the identification of physiological structure in real-world data, a synthetic seed dataset containing 15 athletes and 32 biomarkers was constructed. This seed dataset was not used to learn new clusters, but rather to embed controlled physiological patterns derived from state-of-the-art studies on elite athletes. Complex biomarkers, including insulin and homocysteine, were modeled using normal distributions informed by recent literature on metabolic, hormonal, and cardiovascular responses \cite{saidi2021hematology,Souglis2024,Reale2024}.

The synthetic seed serves as a theoretical ground truth to test whether the unsupervised framework can detect subtle multivariate patterns under controlled conditions.

\subsection{GMM-Based Data Augmentation and Robustness Validation}
For multivariate analysis involving 32 biomarkers, the synthetic seed alone violates the minimum observation-to-variable ratio required for statistical stability. According to \cite{Hair2019}, a ratio of at least 5:1 is necessary to mitigate overfitting, with 10:1 being ideal. To address this constraint and overcome the curse of dimensionality, a Gaussian Mixture Model (GMM) was employed for data augmentation. To ensure mathematical stability given the small seed size, we applied a diagonal covariance matrix (`covariance\_type='diag'`) to reduce parameter estimation complexity, and a regularization factor (`reg\_covar=0.1`) to prevent overfitting.

GMM was used exclusively as a robustness and scalability validation mechanism. Generative models have emerged as a standard approach for addressing data scarcity and privacy constraints in health research \cite{waseem2025review}. The joint probability density function is modeled as presented in Equation \ref{eq3}, where $\Sigma_m$ represents a diagonal covariance matrix with added regularization. This configuration ensures non-singularity, preserves the physiological centroids defined by the seed profiles, and mitigates the curse of dimensionality.
 
\begin{equation}
p(x) = \sum_{m=1}^{M} w_m \mathcal{N}(x \mid \mu_m, \Sigma_m)
\label{eq3}
\end{equation}

The trained GMM generated an expanded cohort of 290 athletes, which was subsequently evaluated using Principal Component Analysis (PCA) to confirm the preservation of cluster topology and spatial organization observed in the real-world data. Importantly, no new physiological profiles were learned from synthetic data; the augmented cohort served solely to validate the clustering solution's structural stability.

A proof-of-concept implementation of the proposed framework was developed in Python using Jupyter Notebooks as an experimental environment to support systematic evaluation across different datasets, clustering strategies, and validation scenarios. The source code is modular and publicly available at \url{https://github.com/FBRosito/unsupervised-athlete-biomarker-clustering}. The implementation supports the full pipeline described in this section, including preprocessing, clinical safety screening, unsupervised clustering, centroid-based interpretation, and robustness validation via synthetic data augmentation. This experimental setup enables reproducible assessment under both real-world and augmented conditions, providing empirical evidence for the feasibility of the proposed architecture as a computational system for athlete monitoring rather than a purely theoretical construct.
\section{Evaluation and Results}
We conduct an empirical evaluation of the proposed framework across real-world and augmented datasets. The analyses focus on assessing clustering stability, physiological coherence, and sensitivity to multivariate patterns under different configurations, rather than predictive performance against injury outcomes. Results are reported with an emphasis on interpretability and robustness, reflecting the intended use of the framework as physiological monitoring and decision support under real-world data constraints. The rest of this section provides details of the evaluations conducted and insights obtained.

\subsection{Algorithmic Validation on Real-World Dataset ($n = 22$)}
The preliminary analysis utilizing real-world amateur data served as a proof-of-concept, stress-testing the pipeline's robustness. Crucially, the Dendrogram-based safety screening automatically flagged two subjects exhibiting extreme Euclidean distances ($>30$ units). Subsequent inspection of the raw data confirmed CK levels exceeding 3,000 U/L, thereby validating the efficacy of the clinical safety module.

\begin{figure*}[ht]
    \centering
    \includegraphics[width=0.9\textwidth]{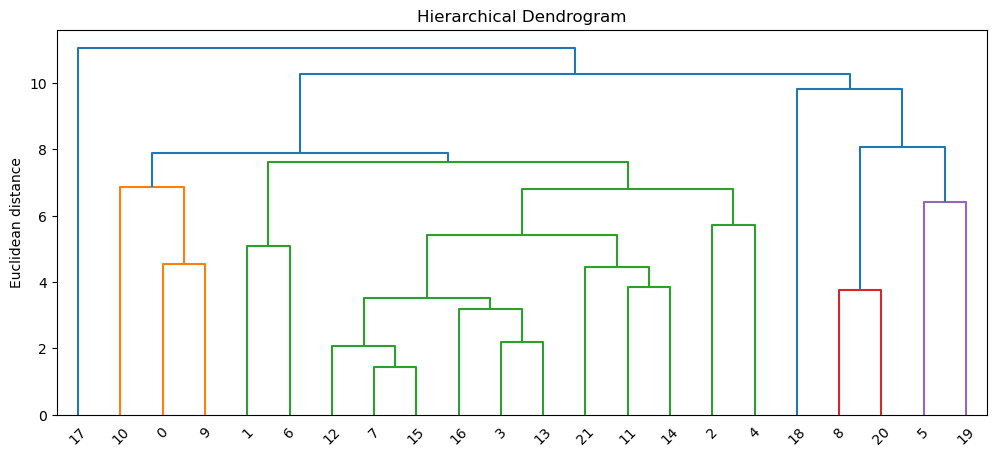}
    \caption{Hierarchical Dendrogram (Ward’s Method) applied to real-world data ($n=22$)}
    \label{fig:hierarchical_dendogram_base_data}
\end{figure*}

Regarding algorithmic selection, Ward’s Hierarchical method demonstrated superior stability compared to K-Means for this limited sample size, effectively mitigating stochastic variation in cluster assignment. Silhouette analysis pinpointed $k=3$ (Score: 0.185) as the optimal cutoff for macro-management, whereas $k=5$ (Score: 0.162) was identified for micro-management, thus establishing the model’s multi-level architecture. This stability is shown in Figure \ref{fig:hierarchical_dendogram_base_data}, where the stability of the tree structure justified the selection of this algorithm over K-Means for the initial small-data phase.

\subsection{Data Augmentation ($n=290$)}
The GMM algorithm successfully expanded the synthetic seed ($n=15$) by generating 275 novel profiles, yielding a total cohort of $n=290$. Table \ref{table:atheletaphysio} details the distribution of the resulting clusters, confirming that the physiological coherence defined during the synthetic generation in the seed phase was preserved.

\begin{table*}[ht]
\centering
\caption{Athlete physiological classification by cluster}
\label{table:atheletaphysio}
\begin{tabular}{c c c c c}
\toprule
\textbf{\makecell{Cluster\\ID}} &
\textbf{\makecell{Physiological\\Classification}} &
\textbf{\makecell{Number of\\Athletes}} &
\textbf{Population (\%)} &
\textbf{\makecell{Biological Signature\\(Mean Z-Score)}} \\
\midrule
3 &
\makecell{Homeostasis\\Maintenance} &
114 &
39.3\% &
\makecell{All vectors oscillating\\within $\pm 0.5\,\sigma$} \\

4 &
\makecell{Anabolic\\Power} &
67 &
23.1\% &
\makecell{Testosterone: $+1.2$\\Cortisol: $-0.8$} \\

1 &
\makecell{Metabolic\\Stress} &
59 &
20.3\% &
\makecell{Cortisol: $+1.8$\\CK: Normal} \\

0 &
\makecell{Mechanical\\Damage} &
37 &
12.7\% &
\makecell{CK: $+2.4$\\LDH: $+2.1$} \\

2 &
Silent Risk &
13 &
4.5\% &
\makecell{Homocysteine: $+2.0$\\Insulin: $+1.5$} \\
\bottomrule
\end{tabular}
\end{table*}

Beyond increasing the effective sample size, the data augmentation process played a critical role in validating the structural stability of the clustering solution. The relative proportions of the identified physiological profiles remained consistent with domain expectations, particularly the predominance of the Homeostasis / Maintenance cluster, which accounted for the largest share of the simulated population. 

\begin{figure*}[ht]
    \centering
    \includegraphics[width=0.80\textwidth]{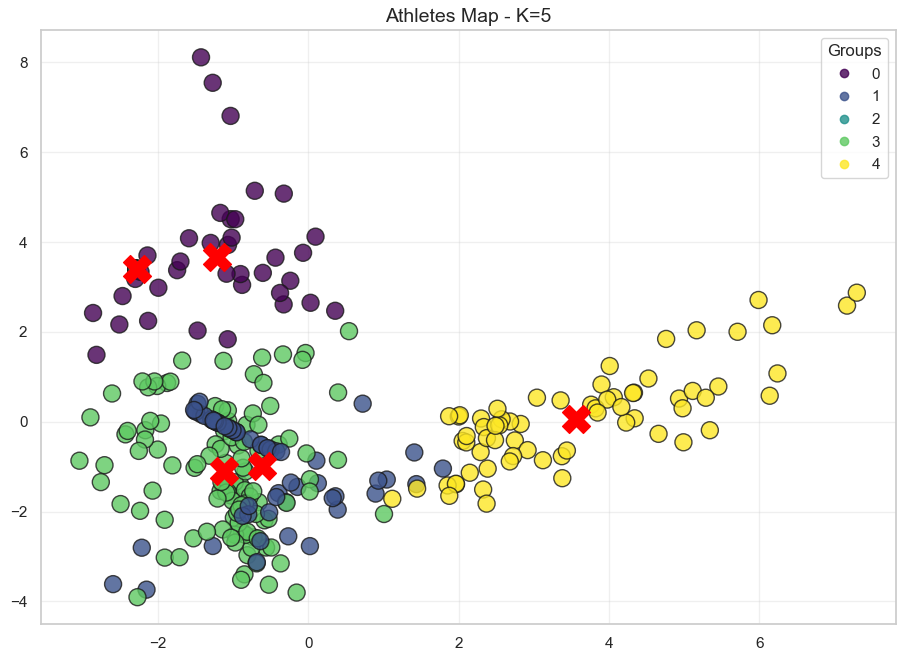}
    \caption{PCA Projection of the expanded synthetic cohort ($n=290$). }
    \label{fig:pca_projection_augmented_extras_data}
\end{figure*}

The PCA projection of the cohort composed of 290 athletes is shown in Figure \ref{fig:pca_projection_augmented_extras_data}. Each color represents a physiological profile. Note the spatial isolation of the silent risk cluster (peripheral points), demonstrating the model's capacity to segregate multivariate outliers.

This outcome indicates that the GMM did not artificially inflate high-risk profiles, but instead preserved a biologically plausible distribution in which the majority of athletes remain in a balanced physiological state. Moreover, the persistence of clear separation between clusters in the augmented space suggests that the identified phenotypes are not artifacts of stochastic variability in the small initial sample, but rather reflect stable multivariate patterns embedded in the biomarker covariance structure. As such, the augmented cohort serves as a robustness check, demonstrating that the proposed framework can scale to larger monitoring scenarios while maintaining physiological coherence and interpretability.

\subsection{Detailed Analysis of Risk Phenotypes}
\begin{figure*}[ht]
    \centering
    \includegraphics[width=1\textwidth]{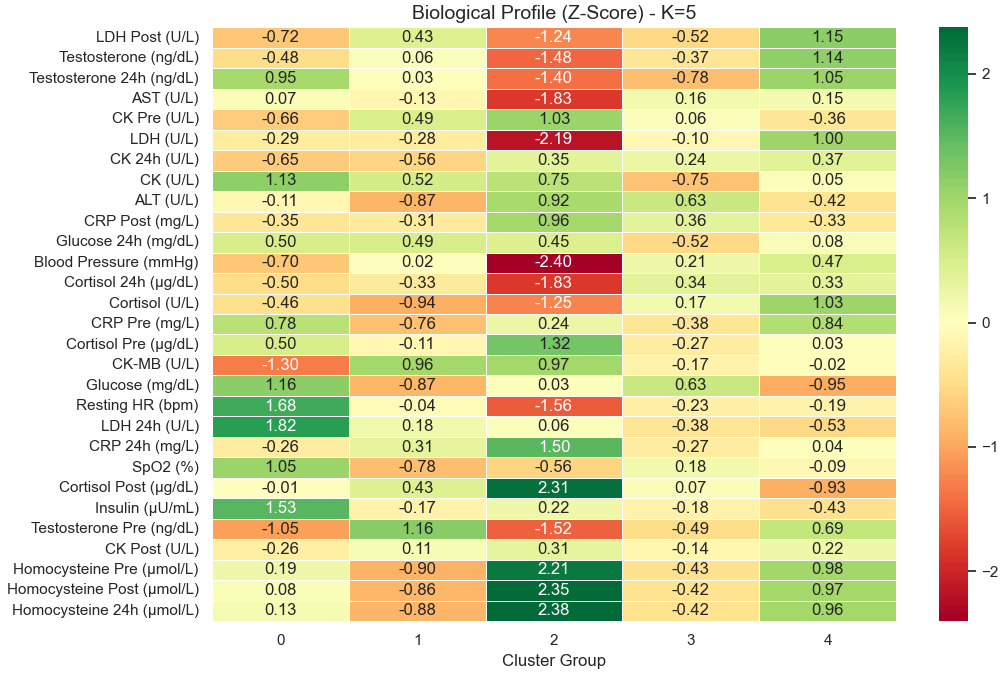}
    \caption{Heatmap of standardized physiological profiles (Z-scores) for the 5 clusters.}
    \label{fig:heatmap_augmented_extras_data}
\end{figure*}

The physiological characterization of the clusters can be visualized through the Z-score heatmap provided in Figure  \ref{fig:heatmap_augmented_extras_data}. The normalization applied revealed distinct biological signatures, where warm colors denote values above the population mean ($Z > 0$) and cool colors denote values below it ($Z < 0$).

Based on this visual distinction, the \(k = 5\) clustering architecture enabled precise disaggregation of fatigue etiologies, overcoming the limitations of traditional binary risk classification. In particular, the model clearly separated athletes experiencing Mechanical Damage (Cluster 0, \(n = 37\)) from those under Metabolic Stress (Cluster 1, \(n = 59\)). The Mechanical Damage profile was characterized by biomarkers associated with peripheral tissue breakdown, whereas the Metabolic Stress profile reflected neuroendocrine axis saturation. This distinction is clinically relevant, as it demonstrates that although approximately 33\% of the synthetic population was classified as ``at risk,'' these athletes required fundamentally different recovery strategies, namely localized physical recovery versus systemic metabolic intervention.

In contrast, the maintenance cluster (Cluster 3, \(n = 114\)) represented athletes in a baseline physiological state, with biomarkers oscillating around homeostatic levels. The predominance of this cluster, accounting for 39.3\% of the simulated population, indicates that the generative process preserved an essential biological property of athlete monitoring: the majority of a healthy squad should remain in homeostasis at any given time. This outcome is particularly important, as it suggests that the proposed framework avoids excessive false positive risk classifications while maintaining sensitivity to meaningful physiological deviations.

\subsection{Algorithmic Sensitivity Validation}
To evaluate the framework's sensitivity to subtle anomalies, a latent silent risk profile was embedded during the synthetic seed construction (Phase II). The unsupervised learning algorithm successfully identified this distinct cluster (Cluster 2, $n = 13$), isolating multivariate patterns of insulin and homocysteine covariance that would remain undetected under conventional monitoring. Athletes assigned to this cluster exhibited normal values of commonly used markers such as creatine kinase and cortisol and would therefore be classified as fit to play under traditional univariate assessment.

Despite the apparent absence of overt stress signals, multivariate analysis revealed an abnormal covariance pattern between homocysteine and insulin. This combination suggests the presence of a latent metabolic and cardiovascular stress signature that is not captured by isolated biomarker thresholds. Importantly, this profile emerged directly from the data structure, without predefined labels or risk annotations.

This result highlights the \emph{algorithmic sensitivity} of the proposed architecture. The unsupervised clustering process successfully isolated a subtle, embedded risk pattern solely from multivariate relationships, demonstrating the system’s ability to detect physiologically meaningful anomalies that are effectively invisible to reductionist approaches. As such, the identification of the ``Silent Risk'' phenotype illustrates the potential of the framework to support proactive surveillance and early intervention, particularly in scenarios where clinical symptoms or performance decrements have not yet manifested.

\section{Discussion and Limitations}

The following discussion examines the implications of the proposed framework from complementary perspectives relevant to healthcare informatics and applied sports monitoring. In particular, we analyze how unsupervised multivariate profiling addresses limitations of binary risk stratification, supports interpretation under real-world constraints on biomarker acquisition, and provides robustness against small-sample effects. Together, these aspects situate the contribution not as a predictive injury model, but as an interpretable decision-support framework for physiological monitoring under operational and clinical constraints.

\subsection{Overcoming Risk Dichotomy}
The seminal contribution by Rossi et al.\ (2023) established that biomarker clustering holds predictive power for injury risk. However, by constraining the analysis to a binary framework (\(k = 2\)), the current ``gold standard'' sacrifices etiological specificity for statistical sensitivity. Our findings demonstrate that the ``High Risk'' category is not a homogeneous condition, but rather an aggregation of distinct physiological phenotypes driven by different underlying stress mechanisms.

Simulation results obtained at higher clustering resolution (\(k = 5\)) reveal that the Metabolic Stress cluster, dominated by hypercortisolism, and the Mechanical Damage cluster, dominated by creatine kinase elevation, occupy opposing regions of the multivariate biomarker space. This distinction is not merely descriptive, but clinically meaningful, as these phenotypes demand fundamentally different recovery strategies. From a data engineering perspective, reliance on extensive historical injury labels, as adopted in supervised approaches such as Rossi’s model, introduces a ``cold start'' problem for new clubs or developing teams. In contrast, the proposed unsupervised framework enables the initiation of a monitoring system based solely on biomarker structure, supporting early and etiology-driven interventions even in the absence of consolidated injury databases.

\subsection{Multivariate Modeling Under Real-World Biomarker Constraints}

A critical challenge in elite sports monitoring is the practical difficulty of acquiring biomarkers. Blood sampling is invasive, costly, and logistically demanding, which explains why most professional environments rely on a limited subset of biomarkers, often restricted to 2 or 3 markers, such as creatine kinase and cortisol. While advances in laboratory technologies allow the measurement of dozens of biomarkers, operational constraints prevent their routine adoption in real-world settings.

This work does not assume unlimited data availability. Instead, it addresses a complementary and often overlooked problem: as the number of measurable biomarkers increases, the complexity of their interactions quickly exceeds human interpretative capacity. Correlations between biomarkers are non-trivial, frequently non-linear, and highly context-dependent. As a result, even when richer biomarker panels are available, physiologists and analysts face a substantial cognitive burden when attempting to extract actionable insight from high-dimensional data.

The proposed framework directly tackles this issue by shifting complexity from human interpretation to computational modeling. Rather than requiring practitioners to reason over dozens of interacting variables, the system transforms multivariate biomarker patterns into a small number of physiologically interpretable states. In this sense, the contribution of the framework is not to increase data demands but to enable principled interpretation when multivariate data are available, while remaining adaptable to reduced biomarker panels.

\subsection{GMM for Data Generation in Sport}
A recurring critique in sports science literature, emphasized by Foucaud et al.\ (2025), concerns the difficulty of statistical generalization due to structurally small sample sizes (\(n < 20\)). In this context, GMMs were not used to artificially inflate datasets but rather as a robustness-validation mechanism, employing diagonal covariance constraints to mitigate overfitting in small-sample regimes.

By demonstrating that cluster topology remains stable across an expanded synthetic population of 290 subjects, we provide evidence that the identified physiological profiles are not artifacts of overfitting to a limited sample. Moreover, the initial training on amateur athlete data, which are intrinsically noisier and more heterogeneous than elite datasets, provides the framework with high ecological validity. If cohesive, physiologically interpretable clusters can be extracted under such noisy conditions, the framework is mathematically well positioned to perform at least as well in more controlled elite environments.

\subsection{Validation of Algorithmic Sensitivity and Latent Risk Detection}

The automatic detection of the silent risk cluster (Cluster 2, \(n = 13\)) further confirms the proposed framework's sensitivity. Although this profile was synthetically introduced during the seed phase as a ground truth pattern, the unsupervised algorithm successfully isolated it without prior labeling. This demonstrates the method’s ability to detect subtle multivariate anomalies that remain invisible under traditional univariate or threshold-based monitoring.

Importantly, athletes in this cluster would be deemed fit to play under conventional assessment criteria, with normal creatine kinase and cortisol levels. The identification of abnormal covariance between homocysteine and insulin shows how unsupervised multivariate analysis can support proactive surveillance by identifying latent metabolic and cardiovascular stress signatures before overt clinical manifestations emerge. This abnormal covariance was modeled in the synthetic dataset and serves as a proof-of-concept of the proposed framework's capacity to identify behaviors and more complex correlations.

\subsection{Decision-Support Implications}
The proposed framework is designed to serve as a decision-support component within athlete-monitoring and health information systems. Instead of producing binary injury-risk predictions, the system provides interpretable physiological-state assignments based on multivariate biomarker structure. Each athlete observation is associated with a cluster characterized by centroid profiles, enabling practitioners to distinguish between physiological conditions such as metabolic stress, mechanical damage, and stable homeostasis.

This representation allows clinicians, sports scientists, and performance staff to identify atypical physiological responses early and adjust training loads, recovery protocols, or clinical assessments accordingly. Because the framework operates without requiring injury labels, it can be deployed in real-world monitoring scenarios where labeled outcomes are unavailable or delayed. Furthermore, the centroid-based interpretation supports transparent reasoning, enabling domain experts to validate model outputs against physiological knowledge.

In practice, the framework can be integrated into digital health and athlete-monitoring platforms that collect biomarker measurements periodically. The system can automatically assign physiological states, track temporal transitions, and provide actionable insights to support individualized training management and injury prevention strategies. This positions the framework as an interpretable, scalable decision-support tool suitable for integration into health information systems and athlete-monitoring platforms.

\subsection{Limitations}
This study should be interpreted in light of the simulated nature of the data expansion phase. Although regularized GMM-based generation preserves the core physiological profiles, the actual prevalence and clinical relevance of the Silent Risk phenotype, estimated at 4.5\% in our simulations, require confirmation in longitudinal observational cohorts. Furthermore, the absence of prospective injury ground truth in the amateur dataset restricts validation to physiological constructs, precluding the computation of predictive accuracy metrics such as the F1 score reported in supervised studies.

Additional work should integrate this clustering pipeline with external load metrics, GPS data, and medical records to close the loop between internal physiological profiling and clinical outcomes. Additionally, systematic evaluation of model performance under progressively reduced biomarker panels will be essential to quantify trade-offs between invasiveness, operational feasibility, and analytical sensitivity in elite sports environments.

\section{Conclusions and Future Work}
This work introduces an unsupervised computational framework for multivariate analysis of biomarkers to identify physiological states and stratify risk in athletes. By operating directly in the joint biomarker space, the proposed framework overcomes the limitations of univariate and binary risk models, enabling etiological differentiation between mechanical and metabolic stress responses.

The results demonstrate that meaningful physiological phenotypes can be identified even under the small-sample conditions typical of real-world sports environments. The successful identification of a latent silent risk phenotype further highlights the capacity of unsupervised learning to reveal subtle multivariate anomalies that are overlooked by traditional monitoring strategies. Importantly, the framework preserves the biological expectation that most athletes remain in homeostasis, thereby avoiding excessive false-positive risk classification.

From a practical perspective, this work addresses two critical challenges faced by elite sports practitioners: the limited feasibility of extensive biomarker collection and the difficulty of interpreting complex biomarker interactions. By translating high-dimensional data into interpretable physiological states, the proposed framework reduces cognitive burden on physiologists and analysts while retaining clinical relevance.

Overall, this work provides a scalable, interpretable foundation for developing clinical decision support systems for athlete monitoring, bridging the gap between advanced analytics and real-world sports medicine practice. As future work, we envision focusing on three main directions. First, prospective validation using longitudinal datasets will be pursued to evaluate the association between identified physiological states and subsequent injury or performance outcomes. Second, the framework will be extended to integrate external load metrics and subjective measures, enabling a more holistic representation of athlete stress and recovery. Finally, adaptive strategies for operating with reduced biomarker panels will be explored to maintain model sensitivity while minimizing invasiveness and operational costs.

\section*{Acknowledgements}
This work is part of the StartHealth project, registered under the number 1273/2025, at the Federal University of Health Sciences of Porto Alegre (UFCSPA), and is supported by Levino Inova startup.

\section*{Authors Contributions}
\textbf{Conceptualization, Methodology, Formal Analysis and Investigation, Writing - original draft preparation:} Fernando Barcelos Rosito and Muriel Figueredo Franco; \textbf{Formal Analysis and Investigation:} Sebastião De Jesus Menezes; \textbf{Resources:} Adriana Seixas, Sebastião De Jesus Menezes, and Simone Sturza; \textbf{Supervision:} Muriel Figueredo Franco. All authors read, commented, and approved the final manuscript.

\section*{Data Availability}
\blue{All data generated are publicly available at https://github.com/FBRosito/unsupervised-athlete-biomarker-clustering under an MIT license.}

\section*{Funding Declaration}
This research received no specific grant from any funding agency in the public, commercial, or not-for-profit sectors.

\balance
\bibliographystyle{IEEETranCustomized}
\bibliography{references.bib}

@article{saidi2021hematology,
  title={Hematology, hormones, inflammation, and muscle damage in elite and professional soccer players: a systematic review with implications for exercise},
  author={Saidi, Karim and Abderrahman, Abderraouf Ben and Hackney, Anthony C and Bideau, Benoit and Zouita, Sghaeir and Granacher, Urs and Zouhal, Hassane},
  journal={Sports medicine},
  volume={51},
  number={12},
  pages={2607--2627},
  year={2021},
  publisher={Springer},
  doi     = {10.3390/ijerph21111394},
  url     = {https://doi.org/10.3390/ijerph21111394}
}

@article{Rossi2023,
  author  = {Rossi, A. and Pappalardo, L. and Filetti, C. and Cintia, P.},
  title   = {Blood sample profile helps to injury forecasting in elite soccer players},
  journal = {Sport Sciences for Health},
  year    = {2023},
  volume  = {19},
  number  = {1},
  pages   = {285--296},
  doi     = {10.1007/s11332-022-00932-1},
  url     = {https://doi.org/10.1007/s11332-022-00932-1}
}

@article{Foucaud2025,
  author  = {Foucaud, A. and Durand, F. and Meric, H.},
  title   = {Using unsupervised machine learning to characterize recovery patterns in elite canoe-kayak athletes across the {Olympic} training year},
  journal = {Frontiers in Sports and Active Living},
  year    = {2025},
  volume  = {7},
  pages   = {1629924},
  doi     = {10.3389/fspor.2025.1629924},
  url     = {https://doi.org/10.3389/fspor.2025.1629924}
}

@article{Souglis2024,
  author  = {Souglis, A. G. and Bogdanis, G. C. and Chryssanthopoulos, C. and Apostolidis, N. and Geladas, N. D.},
  title   = {Position-specific biomarker responses to match vs. {VAMEVAL} test modalities in elite female soccer players},
  journal = {Cogent Engineering},
  year    = {2024},
  volume  = {11},
  number  = {1},
  pages   = {2331188},
  doi     = {10.1080/23311886.2024.2447399},
  url     = {https://doi.org/10.1080/23311886.2024.2447399}
}

@article{Reale2024,
  author  = {Reale, R. and Slater, G. and Burke, L. M.},
  title   = {Individualised dietary strategies for {Olympic} combat sports: Acute weight loss, recovery and competition nutrition},
  journal = {European Journal of Sport Science},
  year    = {2024},
  volume  = {17},
  number  = {6},
  pages   = {727--740},
   doi     = { 10.1080/17461391.2017.1297489},
  url     = {https://doi.org/10.1080/17461391.2017.1297489}
 
}

@article{Popczyk2025,
  author  = {Popczyk, D.},
  title   = {Classifying Soccer Players Based on Physical Capacities and Match-Specific Running Performance Using Machine Learning},
  journal = {Journal of Sports Science and Medicine},
  year    = {2025},
  volume  = {24},
  pages   = {1--12},
   doi     = {10.52082/jssm.2025.764},
  url     = {https://doi.org/10.52082/jssm.2025.764}
}

@article{waseem2025review,
  title={Review of generative AI for synthetic data generation: a healthcare perspective},
  author={Waseem, Hafiz Muhammad and Islam, Saif Ul and Matragkas, Nikolaos and Epiphaniou, Gregory and Arvanitis, Theodoros N and Maple, Carsten},
  journal={Artificial Intelligence Review},
  year={2025},
  publisher={Springer},
  doi = {10.1007/s10462-025-11440-2},
  url = {https://doi.org/10.1007/s10462-025-11440-2}
}

@book{Hair2019,
  author    = {Hair, Joseph F. and Black, William C. and Babin, Barry J. and Anderson, Rolph E.},
  title     = {Multivariate Data Analysis},
  edition   = {8},
  publisher = {Cengage Learning},
  address   = {Hampshire},
  year      = {2019},
}

@article{berisha2021digital,
  author  = {Berisha, Visar and Krantsevich, Chelsea and Hahn, P. Richard and Hahn, Shira and Dasarathy, Gautam and Turaga, Pavan and Liss, Julie},
  title   = {Digital medicine and the curse of dimensionality},
  journal = {npj Digital Medicine},
  year    = {2021},
  volume  = {4},
  number  = {1},
  pages   = {153},
  doi     = {10.1038/s41746-021-00521-5},
  url     = {https://doi.org/10.1038/s41746-021-00521-5}
}

@article{Contreras2024,
  author  = {Contreras-Díaz, G. and Galiano, A. and García-Manso, J. M. and others},
  title   = {The Role of Biomarkers in Monitoring Chronic Fatigue Among Male Professional Team Athletes: A Systematic Review},
  journal = {Sensors},
  year    = {2024},
  volume  = {24},
  pages   = {6862}
}

@article{Haller2023,
  author  = {Haller, N. and Behringer, M. and Reichel, T. and others},
  title   = {Blood-Based Biomarkers for Managing Workload in Athletes: Considerations and Recommendations for Evidence-Based Use of Established Biomarkers},
  journal = {Sports Medicine},
  year    = {2023},
  volume  = {53},
  number  = {7},
  pages   = {1315--1333},
  doi     = {10.1007/s40279-023-01836-x},
  url     = {https://doi.org/10.1007/s40279-023-01836-x}
}

@article{LopezCuervo2025,
  author  = {López-Cuervo, J. M. and Rojas-Jaramillo, A. and García-Caro, A. and González-Santamaria, J. and Humeres, G. and Stout, J. R. and Odriozola-Martínez, A. and Bonilla, D. A.},
  title   = {Biochemical and Perceptual Markers of Physiological Stress During Acute Exercise Overload in U20 Elite Basketball Players},
  journal = {Stresses},
  year    = {2025},
  volume  = {5},
  number  = {3},
  pages   = {52},
  doi     = {10.3390/stresses5030052},
  url     = {https://doi.org/10.3390/stresses5030052}
}

@article{Armstrong2022,
  author  = {Armstrong, L. E. and VanHeest, J. L. and O'Connor, H. and Kraemer, W. J. and Meeusen, R.},
  title   = {Overtraining Syndrome: A Complex Systems Phenomenon},
  journal = {Frontiers in Network Physiology},
  year    = {2022},
  volume  = {1},
  pages   = {794392},
  doi     = {10.3389/fnetp.2022.794392},
  url     = {https://doi.org/10.3389/fnetp.2022.794392}
}

@article{VanEetvelde2021,
  author  = {Van Eetvelde, H. and Mendonça, L. D. and Ley, C. and Seil, R. and Tischer, T.},
  title   = {Machine Learning Methods in Sport Injury Prediction and Prevention: A Systematic Review},
  journal = {Journal of Experimental Orthopaedics},
  year    = {2021},
  volume  = {8},
  number  = {1},
  pages   = {27},
  doi     = {10.1186/s40634-021-00346-x},
  url     = {https://doi.org/10.1186/s40634-021-00346-x}
}

@article{Losciale2024,
  author  = {Losciale, Justin M. and Truong, Linda K. and Ward, Patrick and Collins, Gary S. and Bullock, Garrett S.},
  title   = {Limitations of Separating Athletes into High or Low-Risk Groups based on a Cut-Off: A Clinical Commentary},
  journal = {International Journal of Sports Physical Therapy},
  year    = {2024},
  volume  = {19},
  number  = {9},
  pages   = {1151--1164},
  doi     = {10.26603/001c.122644},
  url     = {https://doi.org/10.26603/001c.122644}
}

@article{Martin2022,
  author  = {Martin, Jack A. and Stiffler-Joachim, Mikel R. and Wille, Christa M. and Heiderscheit, Bryan C.},
  title   = {A Hierarchical Clustering Approach for Examining Potential Risk Factors for Bone Stress Injury in Runners},
  journal = {Journal of Biomechanics},
  year    = {2022},
  volume  = {141},
  pages   = {111136},
  doi     = {10.1016/j.jbiomech.2022.111136},
  url     = {https://doi.org/10.1016/j.jbiomech.2022.111136}
}

@article{Leckey2024,
  author  = {Leckey, Christopher and van Dyk, Nicol and Doherty, Cailbhe and Lawlor, Aonghus and Delahunt, Eamonn},
  title   = {Machine Learning Approaches to Injury Risk Prediction in Sport: A Scoping Review with Evidence Synthesis},
  journal = {British Journal of Sports Medicine},
  year    = {2024},
  volume  = {59},
  number  = {7},
  pages   = {e108576},
  doi     = {10.1136/bjsports-2024-108576},
  url     = {https://doi.org/10.1136/bjsports-2024-108576}
}

@article{pedlar2019blood,
  author  = {Pedlar, Charles R. and Newell, John and Lewis, Nathan A.},
  title   = {Blood biomarker profiling and monitoring for high-performance physiology and nutrition: current perspectives, limitations and recommendations},
  journal = {Sports Medicine},
  year    = {2019},
  volume  = {49},
  number  = {Suppl 2},
  pages   = {185--198},
  doi     = {10.1007/s40279-019-01158-x},
  url     = {https://doi.org/10.1007/s40279-019-01158-x}
}

@article{gonzalez2025unsupervised,
  author  = {González-Martos, Raquel and Galeano, Javier and Ramirez-Castillejo, Carmen and Gusi, Narcis and Gesteiro, Eva and Vicente-Rodriguez, German and Ara, Ignacio and Guadalupe-Grau, Amelia},
  title   = {Unsupervised clustering of biochemical markers reveals health profiles associated with function and survival in active aging},
  journal = {Scientific Reports},
  year    = {2025},
  volume  = {15},
  number  = {1},
  pages   = {30546},
  doi     = {10.1038/s41598-025-14580-1},
  url     = {https://doi.org/10.1038/s41598-025-14580-1}
}

@article{van2024synthetic,
  author  = {van Breugel, Boris and Liu, Tennison and Oglic, Dino and van der Schaar, Mihaela},
  title   = {Synthetic data in biomedicine via generative artificial intelligence},
  journal = {Nature Reviews Bioengineering},
  year    = {2024},
  volume  = {2},
  number  = {12},
  pages   = {991--1004},
  doi     = {10.1038/s44222-024-00245-7},
  url     = {https://doi.org/10.1038/s44222-024-00245-7}
}
\newpage

\end{document}